\documentclass[pdflatex,sn-mathphys-num]{sn-jnl}

\usepackage{graphicx}%
\usepackage{multirow}%
\usepackage{amsmath,amssymb,amsfonts}%
\usepackage{amsthm}%
\usepackage{mathrsfs}%
\usepackage[title]{appendix}%
\usepackage{xcolor}%
\usepackage{textcomp}%
\usepackage{manyfoot}%
\usepackage{booktabs}%
\usepackage{algorithm}%
\usepackage{algorithmicx}%
\usepackage{algpseudocode}%
\usepackage{listings}%

\theoremstyle{thmstyleone}

\theoremstyle{thmstyletwo}%

\theoremstyle{thmstylethree}%

\begin{document}

\title[Article Title]{Nationality and Region Prediction from Names: A Comparative Study of Neural Models and Large Language Models}

\author*[1,2]{\fnm{Keito} \sur{Inoshita}}\email{inosita.2865@gmail.com}

\affil*[1]{\orgdiv{Faculty of Business and Commerce}, \orgname{Kansai University}, \orgaddress{\street{3-3-35 Yamatecho}, \city{Suita}, \postcode{5648680}, \state{Osaka}, \country{Japan}}}

\affil[2]{\orgdiv{Data Science and AI Innovation Research Promotion Center}, \orgname{Shiga University}, \orgaddress{\street{1-1-1 Baba}, \city{Hikone}, \postcode{5228522}, \state{Shiga}, \country{Japan}}}

\abstract{Predicting nationality from personal names has practical value in marketing, demographic research, and genealogical studies. Conventional neural models learn statistical correspondences between names and nationalities from task-specific training data, posing challenges in generalizing to low-frequency nationalities and distinguishing similar nationalities within the same region. Large language models (LLMs) have the potential to address these challenges by leveraging world knowledge acquired during pre-training. In this study, we comprehensively compare neural models and LLMs on nationality prediction, evaluating six neural models and six LLM prompting strategies across three granularity levels (nationality, region, and continent), with frequency-based stratified analysis and error analysis. Results show that LLMs outperform neural models at all granularity levels, with the gap narrowing as granularity becomes coarser. Simple machine learning methods exhibit the highest frequency robustness, while pre-trained models and LLMs show degradation for low-frequency nationalities. Error analysis reveals that LLMs tend to make ``near-miss'' errors, predicting the correct region even when nationality is incorrect, whereas neural models exhibit more cross-regional errors and bias toward high-frequency classes. These findings indicate that LLM superiority stems from world knowledge, model selection should consider required granularity, and evaluation should account for error quality beyond accuracy.

}


\maketitle

\section{Introduction}
Personal names serve as an important source of information reflecting an individual's cultural and geographical background \cite{1}. The structure, phonological patterns, and suffixes of names often exhibit characteristics specific to particular languages and regions. Such correspondences between names and nationalities or regions reflect cultural and linguistic features formed over long historical periods, making it theoretically possible to estimate nationality or region from personal names \cite{2}. The nationality and region prediction task holds practical value across various application domains. In marketing, estimating customers' nationalities or cultural backgrounds from their names enables target audience analysis and personalized service delivery \cite{3}. In demographic research, name-based methods are employed to estimate ethnic compositions when nationality information is missing from census or administrative data \cite{4}. In genealogical research and ancestry tracing, names of ancestors provide clues for estimating their regions of origin.

However, the nationality prediction task involves inherent difficulties. First, the same name may be used across multiple countries and regions. For example, ``Lee'' is widely used in Korea, China, Vietnam, and English-speaking countries, making it difficult to uniquely identify nationality from the name alone. Second, due to immigration and international marriages, names increasingly fail to reflect current nationality. Third, distinguishing nationalities within the same linguistic sphere is extremely challenging. For instance, distinguishing Argentine from Uruguayan, or Belarusian from Ukrainian based solely on names is not easy even for humans. Due to these inherent ambiguities, nationality prediction is a task where achieving 100\% accuracy is fundamentally difficult.

Conventionally, machine learning and neural network-based approaches have been widely employed for nationality prediction tasks. Early research proposed logistic regression models using character-level n-gram features \cite{5}. These methods achieved reasonable prediction performance by capturing statistical features of character patterns in names. Subsequently, with the advancement of neural networks, gradient boosting methods utilizing character n-grams \cite{6} and deep learning methods using recurrent neural networks (RNNs) \cite{7} were proposed. More recently, approaches that fine-tune pre-trained language models such as CANINE \cite{8} and XLM-RoBERTa \cite{9} for nationality prediction have emerged. A common characteristic of these conventional methods is that they learn correspondences between names and nationalities from task-specific training data. Models learn statistical correspondences between character patterns and nationality labels in the training data and generalize to unseen names. Within this framework, high performance can be achieved for nationalities with sufficient samples in the training data, while prediction becomes difficult for nationalities with few samples or names with patterns not present in the training data. Furthermore, distinguishing similar nationalities within the same region requires deeper cultural and linguistic knowledge, which conventional methods cannot explicitly leverage.

Recently, large language models (LLMs) have demonstrated revolutionary performance across various natural language processing tasks. State-of-the-art LLMs such as GPT-4, Claude, and Gemini are pre-trained on massive text corpora spanning trillions of tokens and are widely considered to have acquired extensive world knowledge about language, culture, geography, and history \cite{10}. LLMs possess zero-shot learning capabilities that enable them to handle various tasks through prompt design alone, without additional training \cite{11}. This capability may also be effective for nationality prediction by leveraging knowledge about the relationships between personal names and nationalities or cultures. For example, LLMs are expected to have acquired knowledge through pre-training such as ``-ovich is a patronymic suffix typical of Slavic languages, used in Russia, Ukraine, Serbia, and other countries,'' and to leverage this knowledge for nationality prediction. However, no systematic evaluation has been conducted regarding how much LLMs' nationality prediction capabilities surpass those of neural models or what characteristics they exhibit. It remains unclear whether LLMs truly make predictions by leveraging world knowledge or merely perform superficial pattern matching. It is also unknown whether LLM predictions are robust for low-frequency nationalities or exhibit biases dependent on the distribution of pre-training data. Furthermore, the nature of LLM errors and how they differ from those of neural models has not been investigated. Answering these questions is important for understanding the capabilities and limitations of LLMs and provides guidance for appropriate model selection in practical applications.

In this study, we conduct a comprehensive comparison of neural models and LLMs on the nationality and region prediction task. For neural models, we comprehensively evaluate diverse approaches ranging from traditional machine learning methods with manually designed features, to deep learning methods with automatically learned features, to language models pre-trained on large corpora. This allows us to clarify how model complexity and the presence of prior knowledge affect prediction performance. For LLMs, we compare diverse prompting strategies ranging from simple methods that provide only task descriptions, to methods that provide examples, methods that elicit explicit reasoning processes, methods that integrate multiple inference results, and methods that perform self-correction. This allows us to explore guidelines for prompt design that maximize LLM capabilities. For evaluation, we set three levels of prediction granularity (nationality, region, and continent) and analyze the relationship between prediction difficulty and model characteristics. Fine-grained prediction requires the ability to distinguish similar nationalities within the same region, while coarse-grained prediction evaluates the ability to capture broad geographical directions. Through this hierarchical evaluation, we quantify cases where incorrect fine-grained predictions are correct at coarser granularities, distinguishing whether a model is completely wrong or capturing the general direction. Furthermore, we conduct frequency-based stratified analysis to examine how robust models are for low-frequency nationalities and whether frequency-dependent biases exist. Additionally, through qualitative analysis of errors, we reveal differences in error patterns between the two approaches. Through these multifaceted analyses, we elucidate the characteristics, strengths, and weaknesses of neural models and LLMs in detail.

The main contributions of this study are as follows:

\begin{itemize}
    \item[i)] This study systematically compares diverse neural models and LLM prompting strategies on the nationality and region prediction task, demonstrating that LLMs significantly outperform neural models, and showing that this performance gap is attributable to world knowledge acquired during pre-training.
    \item[ii)] Evaluation at three levels of prediction granularity (nationality, region, and continent) reveals that the performance gap between LLMs and neural models narrows as granularity becomes coarser, providing guidance for model selection based on required prediction granularity and showing that LLM world knowledge is effective for fine-grained prediction while character patterns alone suffice for coarse-grained prediction.
    \item[iii)] Error analysis reveals that LLMs tend to make ``near-miss'' errors where they predict the correct region even when the nationality is incorrect, whereas neural models exhibit more pronounced cross-regional errors and prediction bias toward high-frequency classes, demonstrating the importance of considering not only accuracy but also the quality of errors in model evaluation.
\end{itemize}

The remainder of this paper is organized as follows. Section 2 reviews existing research on nationality prediction and related tasks. Section 3 describes the task formulation, models used, and evaluation metrics. Section 4 reports experimental settings and results, and presents multifaceted analyses. Section 5 discusses the implications of our findings and the limitations of this study. Section 6 concludes the paper.

\section{Related Work}
Research on predicting attributes such as nationality and ethnicity from personal names has been pursued for many years across the fields of social science, demography, and information science. This section provides an overview of the development of existing research, from statistical methods to neural network-based approaches.

\subsection{Name-based Attribute Prediction}
Early research primarily employed statistical and rule-based methods. Mateos \cite{12} comprehensively reviewed 13 studies on name-based ethnic classification, reporting sensitivity ranges of 0.67--0.95 and specificity ranges of 0.8--1.0. Treeratpituk and Giles \cite{5} constructed an ethnic classification model using multinomial logistic regression from Wikipedia data, achieving 85\% accuracy. Voicu \cite{13} proposed the BIFSG model that utilizes first names in addition to surnames and geographic information, demonstrating superior accuracy compared to the conventional BISG method. These studies demonstrated that statistical features contained in names are effective for estimating ethnicity and nationality. However, these methods required manual feature engineering and had limitations in capturing complex patterns.

With the advancement of neural networks, deep learning methods were introduced to name-based attribute prediction research. Ye et al. \cite{14} proposed a model using name embeddings learned from 57 million contact lists, achieving an F1 score of 0.795 for classification across 39 nationalities. Lee et al. \cite{15} proposed a method that directly predicts nationality from name strings using RNNs and evaluated it on Olympic athlete data. Hur \cite{16} classified Malaysian multi-ethnic names using LSTM and character embeddings, achieving up to 7\% accuracy improvement on average over conventional methods. Xie \cite{7} developed the rethnicity package using Bi-LSTM models, improving prediction accuracy for minority ethnicities in U.S. voter data. Wolmer and Bezerra \cite{6} proposed a novel ensemble model integrating n-grams and graph structures from data on one million Brazilian immigrants. These studies demonstrated that neural networks can automatically learn complex patterns in names, but achieving sufficient performance requires large amounts of labeled data, and generalization to patterns not present in the training data remained a challenge.

Name-based attribute prediction has been applied in various domains. Kandt and Longley \cite{4} estimated ethnic tendencies based on surname and first name combinations from UK census data and applied this to quantifying social integration. Bessudnov et al. \cite{17} classified 24 ethnicities with 0.82 accuracy using Russian social network data and applied this to analyzing ethnic structures online. Lee and Velez achieved high-accuracy estimation of politicians' ethnicity and race by combining names and images. Meanwhile, concerns about the social impact of these technologies have also been raised. Hafner et al. \cite{18} conducted fairness audits of name-based ethnic classifiers and proposed fairness-focused models for bias removal. Sebo evaluated the performance of the commercial tool NamSor, reporting error rates of 19.5\% for nationality prediction and 12.6\% for continent classification.

Recently, research on name-based attribute estimation using LLMs has also emerged. An et al. \cite{19} demonstrated that LLMs exhibit race, ethnicity, and gender bias in hiring decisions through name manipulation. Sakunkoo and Sakunkoo \cite{20} analyzed the tendency of LLMs to infer social status, race, and gender from names, visualizing implicit hierarchical biases. These studies suggest that LLMs possess knowledge about the relationships between names and attributes, but no research has systematically compared LLMs' nationality prediction capabilities with conventional neural models. This study aims to fill this gap.

\subsection{LLMs and Prompting Strategies}
The emergence of LLMs has brought a major paradigm shift to natural language processing research. Brown et al. \cite{21} released GPT-3 and demonstrated that a large-scale transformer model with 175B parameters can execute new tasks from natural language instructions without task-specific training. This research showed that strong performance can be achieved even in few-shot settings (with only a few examples provided) and established the paradigm of ``prompt as program.'' The discovery of in-context learning capabilities, where models understand tasks from input context and generalize, became the starting point for subsequent prompting research \cite{22}.

Various methods have been proposed to improve the reasoning capabilities of LLMs by making the reasoning process explicit. Kojima et al. \cite{23} showed that LLMs can generate sequential reasoning processes with a single phrase ``Let's think step by step,'' and Wei et al. \cite{24} proposed Chain-of-Thought prompting, where several demonstrations of chains of thought are provided. These methods achieved significant accuracy improvements on mathematical reasoning benchmarks and introduced the concept of LLMs self-generating traces of thought. Wang et al. \cite{25} proposed Self-Consistency, which generates multiple reasoning paths and determines the answer by majority vote to address the variability in Chain-of-Thought reasoning, achieving 10--17\% performance improvement over Chain-of-Thought alone. Zhou et al. \cite{26} proposed Least-to-Most prompting, which solves complex problems by sequentially addressing easier sub-problems first, evolving Chain-of-Thought from static chains of thought to dynamic thought construction processes.

Self-reflective methods where LLMs evaluate and improve their own outputs have also been developed. Shinn et al. \cite{27} proposed the Reflexion framework, where LLMs self-evaluate their outputs and improve through self-reflection, achieving up to 30\% performance improvement through self-improvement. Madaan et al. \cite{28} proposed Self-Refine, which detects and corrects erroneous or ambiguous reasoning through a three-step process of generation, evaluation, and regeneration, establishing a framework that can generically implement reflection loops. These studies introduced metacognitive loops to LLMs and demonstrated new directions for reasoning that leverage self-correction capabilities.

These prompting methods have been primarily evaluated on mathematical and logical reasoning tasks, but their application to knowledge-intensive classification tasks such as predicting nationality from names has not been sufficiently explored. In nationality prediction tasks, how to elicit the cultural and geographical knowledge that LLMs have acquired during pre-training is important, and the effectiveness of each prompting method may depend on task characteristics. In this study, we apply six prompting methods (Zero-shot, Few-shot, Chain-of-Thought, Self-Consistency, Least-to-Most, and Self-Reflection) to the nationality prediction task and systematically evaluate their effectiveness.

\section{Task Definition and Evaluation Framework}
Personal names serve as an important source of information reflecting an individual's cultural and geographical background. The structure and phonological patterns of names often exhibit characteristics specific to particular languages and regions, making it possible to estimate nationality or region of origin from these features. This section describes the formulation of the nationality and region prediction task addressed in this study, the neural models and LLM-based methods to be compared, and the evaluation metrics.

\subsection{Task Definition}
The nationality prediction task is defined as a multi-class classification problem that takes a personal name as input and predicts nationality. Formally, given a name $x = (c_1, c_2, \ldots, c_n)$ as a sequence of characters, the task is to predict a nationality label $y \in Y = \{y_1, y_2, \ldots, y_C\}$, where $C$ represents the total number of nationality classes.

\begin{figure}[t]
    \centering
    \includegraphics[width=\linewidth]{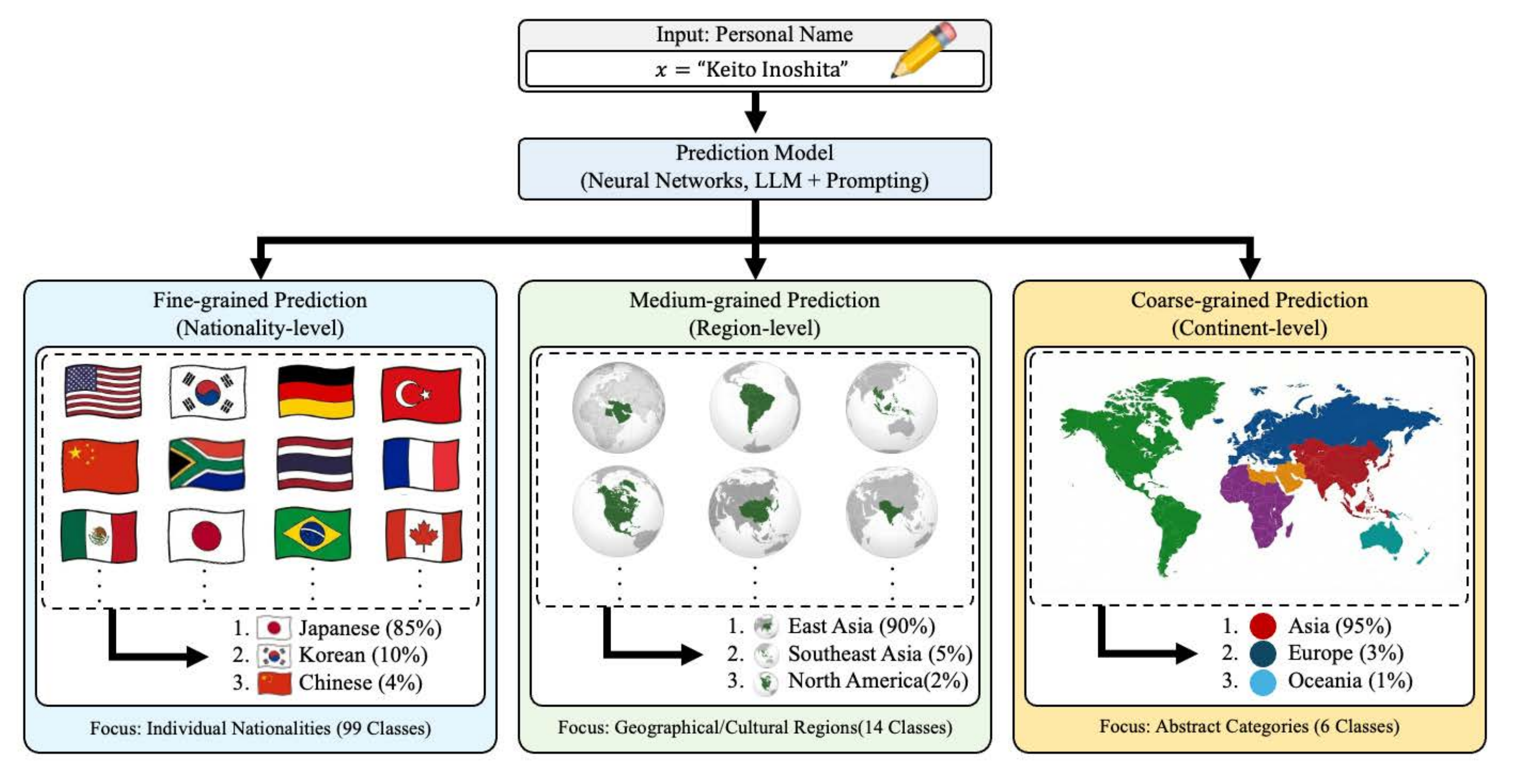}
    \caption{Overview of the nationality and region prediction task. Given a personal name as input, the model predicts nationality at three levels of granularity: fine-grained (99 nationalities), medium-grained (14 regions), and coarse-grained (6 continents).}
    \label{fig:task}
\end{figure}

Nationality prediction varies in difficulty and practical utility depending on the granularity of the prediction target. In this study, we define the following three levels of granularity. Additionally, Fig.~\ref{fig:task} illustrates the concrete task.

\begin{itemize}
    \item[i)] Fine-grained prediction targets nationality-level classification, focusing on individual nationalities (e.g., Japanese, German).
    \item[ii)] Medium-grained prediction targets region-level classification, focusing on regional categories grouped based on geographical and cultural similarities (e.g., East Asia, Western Europe).
    \item[iii)] Coarse-grained prediction targets continent-level classification, focusing on the most abstract categories (e.g., Asia, Europe).
\end{itemize}

This hierarchical structure has two significant implications. First, it enables analysis of how prediction difficulty changes with granularity. Second, it allows quantification of cases where predictions are incorrect at fine granularity but correct at coarser granularity. For example, if Korean is incorrectly predicted as Japanese, it is still correct as East Asia. This enables evaluation of whether a model is capturing the general direction.

Furthermore, since there are many cases where nationality cannot be uniquely identified from a name alone, this study adopts Top-k prediction rather than single prediction. Models output $k$ nationality candidates ranked by confidence, and evaluation is based on whether the correct answer is included in the top $k$ results.

It should be noted that inherent ambiguity exists in nationality prediction tasks. First, there is polysemy of names, where the same name is used across multiple countries and regions. For example, ``Lee'' is widely used in Korea, China, Vietnam, and English-speaking countries. Second, due to immigration and international marriages, there are cases where names do not reflect the country of origin or current nationality. Third, diversity in romanization methods within the same linguistic sphere can make prediction difficult. Due to these challenges, nationality prediction is inherently a task where achieving 100\% accuracy is difficult, making Top-k prediction and evaluation at coarser granularities important.

\subsection{Models}
In this study, we employ six types of neural baseline methods and six types of LLM prompting methods to compare the performance of neural models and LLMs on the nationality and region prediction task.

\subsubsection{Neural Baselines}
As neural baselines, we adopt six types of models that utilize character-level features. Since personal names are relatively short texts and character-level features are considered more effective than word-level features, all models process inputs with characters as the basic unit.

\begin{itemize}
    \item[i)] SVM: A linear classifier using TF-IDF features of character n-grams. Despite being a simple method, it is known as a strong baseline for text classification. In this study, we use character 1-gram to 4-gram features.
    \item[ii)] fastText: A shallow neural network model that learns embeddings of character n-grams. By utilizing subword information, it can obtain robust representations even for unknown or rare words.
    \item[iii)] CNN: A model that applies convolutional neural networks to character embeddings. By using filters of different sizes, it can capture character patterns of various lengths.
    \item[iv)] BiLSTM: A model that applies bidirectional LSTM to character embeddings. It can obtain representations that consider the context of the entire sequence and integrate information from both the beginning and end of names.
    \item[v)] CANINE \cite{8}: A Transformer model pre-trained at the character level. Since it accepts Unicode characters directly as input without using a tokenizer, it enables robust processing for multilingual text.
    \item[vi)] XLM-RoBERTa \cite{9}: A multilingual Transformer model pre-trained on more than 100 languages. Although it uses subword tokenization, it can handle names from various linguistic spheres due to training on multilingual corpora.
\end{itemize}

These models cover traditional machine learning methods (SVM), shallow neural networks (fastText), deep learning methods (CNN, BiLSTM), and pre-trained models (CANINE, XLM-RoBERTa), enabling comparison of performance across diverse approaches.

\subsubsection{LLM Prompting Methods}
As LLM-based methods, we adopt six types of prompting strategies. These methods aim to elicit LLMs' nationality and region prediction capabilities through prompt design alone, without additional training.

\begin{itemize}
    \item[i)] Zero-shot Prompting \cite{21}: The simplest method that provides only the task description and input, performing prediction without examples. Specifically, the name is input along with an instruction such as ``Predict the most likely nationality from the given name,'' and the LLM makes predictions based solely on knowledge acquired during pre-training. This method serves as a baseline for evaluating LLMs' fundamental nationality and region prediction capabilities.
    \item[ii)] Few-shot Prompting \cite{21}: A method that includes a small number of input-output examples in the prompt in addition to the task description. For example, by providing examples such as ``Tanaka Taro $\rightarrow$ Japanese'' and ``John Smith $\rightarrow$ American,'' the task format and expected output are made explicit. Through examples, LLMs can understand the output format and make predictions by referring to similar patterns.
    \item[iii)] Chain-of-Thought Prompting \cite{24}: A method that has the model output the reasoning process leading to the prediction step by step. First, the model is made to analyze the features of the name (phonological patterns, suffixes, etymology, character combinations, etc.), then reason about which nationality or region these features relate to, and finally output the prediction. For example, the model performs stepwise reasoning such as ``This name has the suffix `-ovich,' which is a typical pattern in Slavic languages. Therefore, Russia or Eastern European countries are likely.''
    \item[iv)] Self-Consistency Prompting \cite{25}: A method that performs multiple inferences for the same input and determines the final prediction by majority vote. LLM outputs are probabilistic and may return different results for the same input. In this method, five independent inferences are executed for the same name, and the most frequently predicted nationality or region is adopted as the final result. This enables more stable predictions without relying on a single inference.
    \item[v)] Least-to-Most Prompting \cite{26}: A method that decomposes complex problems into simpler sub-problems and solves them step by step. In this task, since direct prediction of 99 nationalities is difficult, hierarchical reasoning is performed by first predicting ``Which region (Asia, Europe, etc.) does this name belong to'' and then predicting ``Which nationality is most likely within that region.'' By progressively narrowing down from coarse granularity to fine granularity, more accurate predictions are expected.
    \item[vi)] Self-Reflection Prompting \cite{27}: A method that has the model self-evaluate its initial prediction and make corrections as needed. After making an initial prediction, self-critique is prompted with questions such as ``Is this prediction correct? Are there other possibilities?'' to consider the confidence in the initial prediction and potential errors. If confidence is low or alternative candidates exist, the final prediction is output after reconsideration.
\end{itemize}

These methods cover diverse prompting strategies, from simple prompts (Zero-shot, Few-shot) to explicit reasoning processes (Chain-of-Thought), integration of multiple inferences (Self-Consistency), problem decomposition (Least-to-Most), and self-correction (Self-Reflection).

\subsection{Evaluation Metrics}
In text classification tasks, Accuracy, Precision, Recall, and F1 score are commonly used as evaluation metrics. However, the nationality and region prediction task has characteristics that make it difficult to directly apply these standard evaluation frameworks.

First, there is the problem of class imbalance. In nationality prediction, nationalities with large populations such as American, Chinese, and Indian have many samples, while nationalities of small countries or specific regions have extremely few samples. When evaluating imbalanced data using only Accuracy, the performance on majority classes dominates the results, and the performance on minority classes is not adequately reflected.

Second, there is inherent ambiguity in the task. As described in Section 3.1, since the same name may be used across multiple nationalities, evaluation that considers only a single correct answer may be overly strict. For example, if Korean is predicted for the name ``Lee'' when the correct answer is Chinese, it is difficult to consider this a complete error.

Third, it is necessary to consider the ``nearness'' of predictions. When Korean is incorrectly predicted as Japanese versus Brazilian, the quality of errors differs significantly. The former is an error within the same East Asian region, while the latter is an error to a completely different region. Conventional evaluation metrics cannot distinguish such degrees of error.

Based on these challenges, this study employs the following evaluation metrics:

\begin{itemize}
    \item[i)] Accuracy: The accuracy rate of Top-1 predictions, indicating the proportion where the prediction with highest confidence matches the correct label. This is the most fundamental evaluation metric and is used for comparison with other studies, but interpretation must acknowledge the issues mentioned above.
    \item[ii)] Macro-F1: A metric that calculates the F1 score for each class and takes their simple average. Since it weights performance for each class equally, performance on classes with few samples is also appropriately reflected. For tasks like nationality and region prediction where there is significant imbalance in sample sizes across classes, this enables fairer evaluation than Accuracy. Macro-F1 is calculated as follows:
    \begin{equation}
        \text{Macro-F1} = \frac{1}{C} \sum_{i=1}^{C} \text{F1}_i = \frac{1}{C} \sum_{i=1}^{C} \frac{2 \cdot P_i \cdot R_i}{P_i + R_i}
    \end{equation}
    where $C$ is the total number of classes, and $P_i$ and $R_i$ denote the precision and recall for class $i$, respectively.
    \item[iii)] Precision@k: A metric indicating the proportion where the correct label is included in the Top-k predictions. In this study, we use $k = 2, 3, 5$. Since there are cases where uniquely identifying nationality or region from a name is difficult, performing Top-k evaluation in addition to Top-1 allows assessment of whether the model can include the correct answer among its candidates. Precision@1 is equivalent to Accuracy. Precision@k is defined as follows:
    \begin{equation}
        \text{Precision@}k = \frac{1}{N} \sum_{j=1}^{N} I(y_j \in \text{Top}_k(\hat{y}_j))
    \end{equation}
    where $N$ is the total number of samples, $y_j$ is the correct label for sample $j$, $\text{Top}_k(\hat{y}_j)$ is the set of top $k$ predictions for sample $j$, and $I[\cdot]$ is the indicator function.
\end{itemize}

In addition to these metrics, we conduct frequency-based stratified analysis to evaluate model robustness. Specifically, nationalities are divided into three groups based on their frequency in the training data: Head (high-frequency), Mid (medium-frequency), and Tail (low-frequency), and performance is evaluated separately for each group. This analysis enables verification of how robust models are for rare nationalities and whether frequency-dependent biases exist.

Furthermore, in the evaluation of hierarchical granularity (nationality, region, and continent), we also analyze the proportion of cases where incorrect predictions at fine granularity become correct at coarser granularity. This allows us to distinguish whether a model is completely wrong or capturing the general direction.

\section{Experiments and Analysis}
This section experimentally evaluates the performance of neural models and LLMs on the nationality and region prediction task. We first describe the experimental design, including the dataset and implementation details, followed by the results and analysis of each experiment.

\subsection{Experimental Design}
\subsubsection{Dataset}
In this study, we use a nationality prediction dataset constructed based on the name2nat dataset \cite{29}. The name2nat dataset is a large-scale dataset containing personal names and their nationality labels, covering 890,248 romanized personal names and 173 nationalities.

From the original name2nat dataset, data extraction and preprocessing were performed according to the following procedure. First, to ensure sufficient training samples, only nationalities with 500 or more samples were included, excluding nationalities with extremely few samples. This threshold was set as the minimum number of samples necessary for models to learn the characteristics of each nationality. As a result, the number of target nationalities was reduced from 173 to 99. Next, to mitigate extreme class imbalance, an upper limit of 800 samples was set for each nationality, with excess samples randomly removed. This upper limit was determined to balance preventing majority classes from dominating the training while ensuring sufficient data volume. Finally, the data was split into training, validation, and test sets at an 8:1:1 ratio using stratified sampling. Stratified sampling ensured that each nationality was included in all splits at appropriate proportions, followed by shuffling after the split. As a result of preprocessing, the final dataset contains 75,345 samples and 99 nationality classes. Table~\ref{tab:dataset} shows the dataset statistics.

\begin{table}[t]
    \centering
    \caption{Dataset statistics.}
    \label{tab:dataset}
    \begin{tabular*}{0.3\textwidth}{@{\extracolsep{\fill}}lr}
        \hline
        Split & Samples \\
        \hline
        Train & 60,277 \\
        Dev   & 7,534 \\
        Test  & 7,534 \\
        \hline
        Total & 75,345 \\
        \hline
    \end{tabular*}
\end{table}

\begin{figure}[t]
    \centering
     \includegraphics[width=0.8\linewidth]{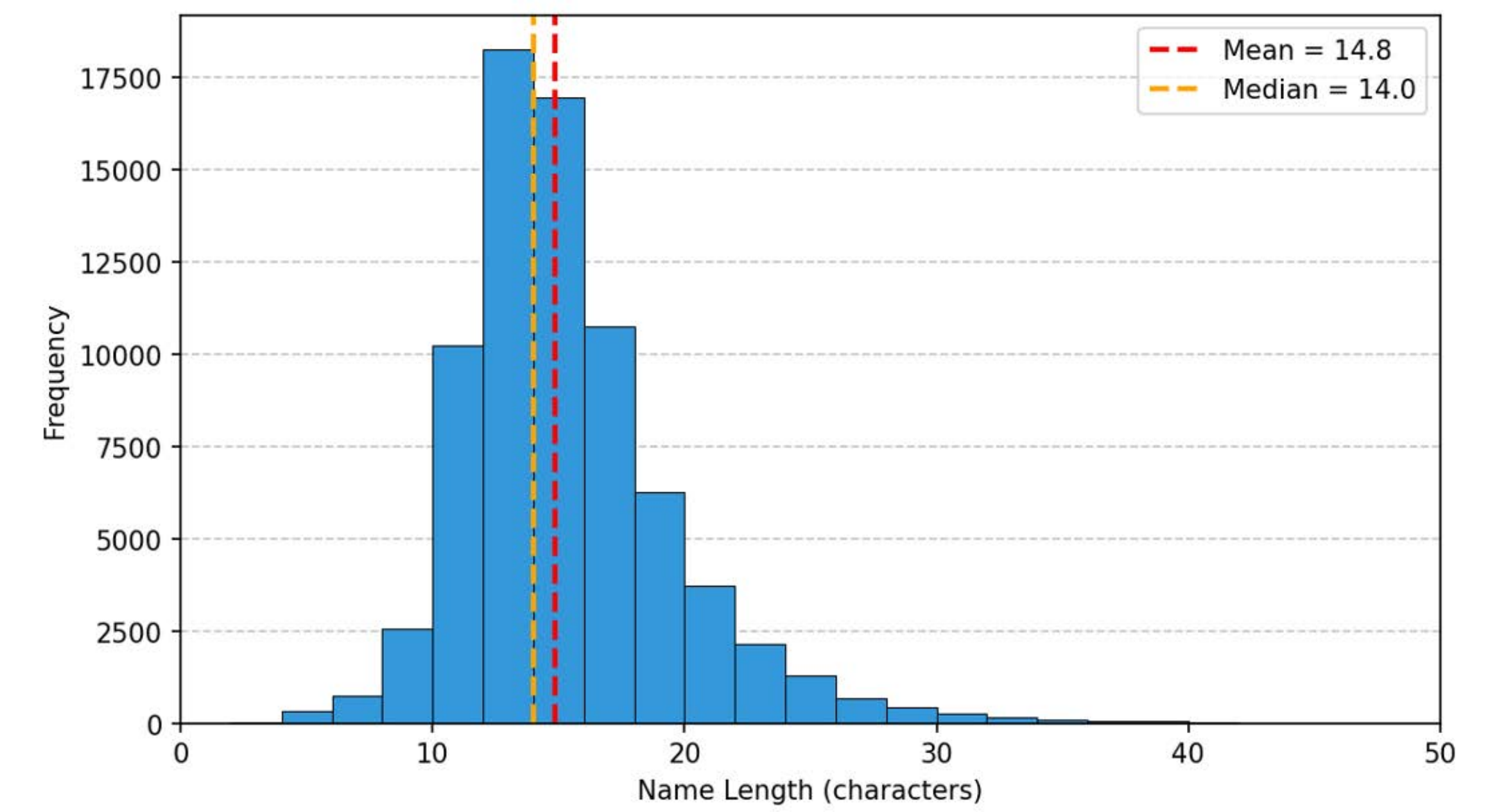}
    \caption{Distribution of name lengths in the dataset. The average length is 14.8 characters with a median of 14.0 characters.}
    \label{fig:name_length}
\end{figure}

Fig.~\ref{fig:name_length} shows the distribution of name lengths in the dataset. The average name length is 14.8 characters with a median of 14.0 characters, and the majority fall within the range of 10 to 20 characters. Extremely short or long names are rare, and models will primarily learn from names of typical length.

For the hierarchical granularity evaluation described in Section 3.1, the 99 nationalities were mapped to 14 regions. The mapping was defined based on geographical location and cultural-linguistic similarities. For example, East Asian nationalities share common characteristics in romanized representations derived from Chinese characters, while Western European nationalities share naming conventions derived from Latin or Germanic languages. Table~\ref{tab:region} shows the distribution of sample counts per region in the training data. There is approximately a 15-fold difference between the most frequent region, Eastern Europe (9,600 samples), and the least frequent region, Northern Europe (640 samples), indicating that imbalance exists not only at the nationality level but also at the region level. This imbalance arises from differences in the number of nationalities included in each region and differences in sample counts for each nationality in the original dataset.

\begin{table}[t]
    \centering
    \caption{Region definitions and statistics.}
    \label{tab:region}
    \begin{tabular}{lrr}
        \hline
        Region &  Nationalities &  Train Samples \\
        \hline
        Eastern Europe & 15 & 9,600 \\
        Africa & 15 & 8,598 \\
        Western Europe & 11 & 7,040 \\
        South America & 10 & 6,285 \\
        Middle East & 10 & 5,837 \\
        Southeast Asia & 7 & 4,480 \\
        South Asia & 6 & 3,801 \\
        Central America \& Caribbean & 7 & 3,625 \\
        Southern Europe & 5 & 3,091 \\
        East Asia & 5 & 2,964 \\
        North America & 3 & 1,920 \\
        Caucasus \& Central Asia & 2 & 1,280 \\
        Oceania & 2 & 1,116 \\
        Northern Europe & 1 & 640 \\
        \hline
    \end{tabular}
\end{table}

The mapping to six continents was defined by further aggregating the 14 regions. Asia includes four regions: East Asia, Southeast Asia, South Asia, and Caucasus \& Central Asia. Europe includes four regions: Western Europe, Northern Europe, Southern Europe, and Eastern Europe. Americas includes three regions: North America, Central America \& Caribbean, and South America. Middle East, Africa, and Oceania were each treated as individual continent categories.

Additionally, for frequency-based stratified analysis, the 99 nationalities were equally divided into three groups based on their frequency in the training data: Head (high-frequency), Mid (medium-frequency), and Tail (low-frequency). Each group contains 33 nationalities, with the Head group containing high-frequency nationalities that reached the upper limit of 800 samples, and the Tail group containing low-frequency nationalities with approximately 500 samples. This division enables evaluation of how robustly models can predict nationalities of different frequencies.

\subsubsection{Implementation Details}
This section describes the implementation details of neural models and LLM prompting methods. All experiments were conducted on an NVIDIA H100 GPU (93GB). Neural models were implemented using Python 3.12, PyTorch 2.7, Hugging Face Transformers 4.57, and scikit-learn 1.6.

All neural models were trained on the same training data, with early stopping based on validation data performance. Table~\ref{tab:hyperparams} shows the key hyperparameters for each model.

\begin{table}[t]
    \centering
    \caption{Hyperparameters for neural models.}
    \label{tab:hyperparams}
    \begin{tabular}{ll}
        \hline
        Model & Key Hyperparameters \\
        \hline
        SVM & n-gram: 1-4, max features: 50,000, C: 1.0 \\
        fastText & n-gram: 2-5, embed dim: 100, lr: 5e-1, epochs: 25 \\
        CNN & max len: 50, embed dim: 64, filters: 128, \\
                 & filter sizes: [2,3,4,5], dropout: 0.5, lr: 1e-3, epochs: 20 \\
        BiLSTM & max len: 50, embed dim: 64, hidden dim: 128, \\
                    & layers: 2, dropout: 0.5, lr: 1e-3, epochs: 20 \\
        CANINE & max len: 64, lr: 2e-5, batch size: 64,  epochs: 10 \\
        XLM-RoBERTa & max len: 32, lr: 2e-5, batch size: 64,  epochs: 10 \\
        \hline
    \end{tabular}
\end{table}

Adam optimizer was used for all neural models except SVM, and AdamW with weight decay of 0.01 was used for pre-trained models (CANINE, XLM-RoBERTa). For deep learning models (Char-CNN, Char-BiLSTM, CANINE, XLM-RoBERTa), early stopping was applied when validation accuracy did not improve for two consecutive epochs. For pre-trained models, google/canine-s and xlm-roberta-base were used. For SVM, Platt Scaling was applied to obtain probability outputs.

For LLMs, GPT-4.1-mini was used through the OpenAI API. The temperature parameter was set to 1.0 to ensure output diversity. The output format was specified as a JSON array, with instructions to return the top 5 nationalities in order of confidence. The maximum output token count was set to 256. Table~\ref{tab:llm_settings} shows the settings for each prompting method.

\begin{table}[t]
    \centering
    \caption{Settings for LLM prompting methods.}
    \label{tab:llm_settings}
    \begin{tabular}{ll}
        \hline
        Method & Settings \\
        \hline
        Zero-shot & Include list of 99 nationalities in system prompt \\
        Few-shot & 14 examples selected from training data (one per region) \\
        Chain-of-Thought & Instruct to output stepwise analysis of name features \\
        Self-Consistency & Execute 5 independent inferences and take majority vote \\
        Least-to-Most & 3-stage inference: 6 continents → 14 regions → 99 nationalities \\
        Self-Reflection & Execute self-evaluation and correction after initial prediction \\
        \hline
    \end{tabular}
\end{table}

API requests to the LLM were parallelized through asynchronous processing, allowing up to 50 concurrent requests. In case of API call failures, up to 3 retries were performed with exponential backoff. Responses from the LLM were parsed in JSON format, and outputs not included in the 99 nationality list or unparseable responses were treated as ``Unknown.'' The Unknown rate was 0.0\% for most methods, and while some Unknown responses were recorded in certain methods, the rate remained below 0.2\%, indicating that the LLM's compliance with the output format was extremely high.

\subsection{Evaluation on Nationality Prediction}
This section compares the performance of neural models and LLMs on the 99 nationality prediction task. All experiments were conducted three times with different random seeds, and the mean and standard deviation are reported. Table~\ref{tab:nationality_results} shows the results.

\begin{table}[t]
    \centering
    \caption{Results on 99 nationality prediction. Values are mean ± standard deviation over 3 runs.}
    \label{tab:nationality_results}
    \begin{tabular}{lcccc}
        \hline
        Method & Accuracy & Macro-F1 & Precision@3 & Precision@5 \\
        \hline
        \multicolumn{5}{c}{Neural Model} \\
        \hline
        SVM & 0.481 ± 0.004 & 0.466 ± 0.005 & 0.644 ± 0.006 & 0.710 ± 0.005 \\
        fastText & 0.253 ± 0.003 & 0.221 ± 0.003 & 0.414 ± 0.004 & 0.505 ± 0.002 \\
        CNN & 0.425 ± 0.009 & 0.407 ± 0.009 & 0.615 ± 0.007 & 0.698 ± 0.005 \\
        BiLSTM & 0.402 ± 0.009 & 0.386 ± 0.011 & 0.600 ± 0.008 & 0.688 ± 0.002 \\
        CANINE & 0.450 ± 0.013 & 0.435 ± 0.013 & 0.648 ± 0.012 & 0.733 ± 0.010 \\
        XLM-RoBERTa & 0.446 ± 0.010 & 0.426 ± 0.009 & 0.647 ± 0.007 & 0.732 ± 0.002 \\
        \hline
        \multicolumn{5}{c}{Large Language Model} \\
        \hline
        Zero-shot & 0.759 ± 0.004 & 0.765 ± 0.004 & 0.853 ± 0.002 & 0.877 ± 0.003 \\
        Few-shot & 0.730 ± 0.005 & 0.731 ± 0.005 & 0.840 ± 0.006 & 0.868 ± 0.005 \\
        Chain-of-Thought & 0.709 ± 0.009 & 0.715 ± 0.009 & 0.821 ± 0.007 & 0.853 ± 0.006 \\
        Self-Consistency & 0.763 ± 0.006 & 0.769 ± 0.005 & 0.855 ± 0.005 & 0.880 ± 0.005 \\
        Least-to-Most & 0.619 ± 0.004 & 0.615 ± 0.004 & 0.722 ± 0.003 & 0.767 ± 0.001 \\
        Self-Reflection & 0.776 ± 0.005 & 0.782 ± 0.004 & 0.870 ± 0.006 & 0.893 ± 0.005 \\
        \hline
    \end{tabular}
\end{table}

Among neural models, SVM achieved the highest performance with Accuracy of 0.481 and Macro-F1 of 0.466. The pre-trained models CANINE and XLM-RoBERTa achieved Accuracy of 0.450 and 0.446 respectively, slightly below SVM. CNN and BiLSTM achieved Accuracy of 0.425 and 0.402, while fastText showed the lowest performance at 0.253. For Precision@5, CANINE and XLM-RoBERTa exceeded 0.730, indicating that differences among neural models tend to narrow in Top-5 prediction accuracy. The reason SVM outperformed pre-trained models may be that direct pattern matching using character n-grams is effective for feature extraction from personal names. On the other hand, pre-trained models may lose character patterns specific to personal names due to subword tokenization splitting the names. The low performance of fastText suggests that shallow architectures cannot sufficiently capture the complex patterns required for nationality prediction.

LLMs significantly outperformed neural models across all prompting methods. Self-Reflection achieved the highest performance with Accuracy of 0.776 and Macro-F1 of 0.782. This represents a difference of 0.295 compared to the best neural model performance (SVM: 0.481). Self-Consistency and Zero-shot also showed high performance with Accuracy of 0.763 and 0.759, respectively. The superiority of LLMs is attributed to the world knowledge acquired through large-scale pre-training. LLMs have learned relationships between personal names and nationalities/cultures during pre-training and can make predictions without depending on task-specific training data. In contrast, neural models learn correspondences between names and nationalities only from the training data for this task, limiting their generalization capability.

Meanwhile, Least-to-Most achieved 0.619, remaining at a lower performance compared to other LLM methods. This method performs inference in three stages (6 continents → 14 regions → 99 nationalities), and error propagation where errors at upper hierarchical levels propagate to lower levels is likely occurring. Interestingly, Few-shot (0.730) and Chain-of-Thought (0.709) performed below Zero-shot (0.759). For Few-shot, the provided examples may have introduced bias toward specific patterns, hindering generalization to diverse names. For Chain-of-Thought, making the reasoning process explicit may have increased the risk of falling into incorrect reasoning paths. These results indicate that advanced prompting methods do not necessarily contribute to performance improvement, suggesting the importance of method selection according to task characteristics.

Focusing on the relationship between Macro-F1 and Accuracy, LLMs show nearly equivalent values for both metrics (e.g., Self-Reflection shows Accuracy of 0.776 and Macro-F1 of 0.782). In contrast, while the difference between both metrics is small for neural models, they are not as closely matched as for LLMs. The fact that Macro-F1 is equal to or higher than Accuracy suggests that LLMs exhibit relatively uniform performance even for minority classes. This point will be analyzed in detail in Section 4.4.

The standard deviation across all methods was approximately 0.01, confirming high reproducibility of results. Notably, LLMs showed stable performance despite the relatively high temperature parameter setting of 1.0.

\subsection{Evaluation on Region Prediction}
This section evaluates performance on the 14-region and 6-region prediction tasks. As described in Section 3.1, evaluation at hierarchical granularity enables quantification of cases where models are incorrect at fine granularity but correct at coarser granularity. Table~\ref{tab:14region_results} shows the results for 14-region prediction, and Table~\ref{tab:6region_results} shows the results for 6-region prediction.

\begin{table}[t]
    \centering
    \caption{Results on 14 region prediction. Values are mean ± standard deviation over 3 runs.}
    \label{tab:14region_results}
    \begin{tabular}{lcccc}
        \hline
        Method & Accuracy & Macro-F1 & Precision@2 & Precision@3 \\
        \hline
        \multicolumn{5}{c}{Neural Model} \\
        \hline
        SVM & 0.682 ± 0.005 & 0.619 ± 0.003 & 0.812 ± 0.004 & 0.872 ± 0.002 \\
        fastText & 0.459 ± 0.000 & 0.360 ± 0.005 & 0.630 ± 0.002 & 0.732 ± 0.004 \\
        CNN & 0.637 ± 0.008 & 0.552 ± 0.015 & 0.787 ± 0.005 & 0.858 ± 0.002 \\
        BiLSTM & 0.647 ± 0.007 & 0.573 ± 0.006 & 0.796 ± 0.008 & 0.865 ± 0.005 \\
        CANINE & 0.683 ± 0.006 & 0.621 ± 0.001 & 0.824 ± 0.003 & 0.888 ± 0.003 \\
        XLM-RoBERTa & 0.690 ± 0.008 & 0.631 ± 0.010 & 0.830 ± 0.002 & 0.894 ± 0.001 \\
        \hline
        \multicolumn{5}{c}{Large Language Model} \\
        \hline
        Zero-shot & 0.715 ± 0.002 & 0.660 ± 0.001 & 0.872 ± 0.002 & 0.932 ± 0.037 \\
        Few-shot & 0.697 ± 0.004 & 0.647 ± 0.006 & 0.869 ± 0.002 & 0.904 ± 0.002 \\
        Chain-of-Thought & 0.697 ± 0.001 & 0.650 ± 0.003 & 0.859 ± 0.001 & 0.905 ± 0.001 \\
        Self-Consistency & 0.710 ± 0.001 & 0.659 ± 0.005 & 0.876 ± 0.001 & 0.914 ± 0.002 \\
        Least-to-Most & 0.661 ± 0.002 & 0.601 ± 0.003 & 0.837 ± 0.002 & 0.877 ± 0.002 \\
        Self-Reflection & 0.751 ± 0.004 & 0.704 ± 0.005 & 0.891 ± 0.001 & 0.919 ± 0.002 \\
        \hline
    \end{tabular}
\end{table}

\begin{table}[t]
    \centering
    \caption{Results on 6 region prediction. Values are mean ± standard deviation over 3 runs.}
    \label{tab:6region_results}
    \begin{tabular}{lcccc}
        \hline
        Method & Accuracy & Macro-F1 & Precision@2 & Precision@3 \\
        \hline
        \multicolumn{5}{c}{Neural Model} \\
        \hline
        SVM & 0.767 ± 0.001 & 0.679 ± 0.007 & 0.899 ± 0.001 & 0.952 ± 0.003 \\
        fastText & 0.573 ± 0.000 & 0.461 ± 0.002 & 0.720 ± 0.128 & 0.895 ± 0.003 \\
        CNN & 0.740 ± 0.005 & 0.632 ± 0.016 & 0.889 ± 0.001 & 0.946 ± 0.002 \\
        BiLSTM & 0.744 ± 0.004 & 0.664 ± 0.004 & 0.890 ± 0.003 & 0.949 ± 0.003 \\
        CANINE & 0.776 ± 0.003 & 0.696 ± 0.009 & 0.906 ± 0.002 & 0.959 ± 0.002 \\
        XLM-RoBERTa & 0.780 ± 0.005 & 0.692 ± 0.012 & 0.911 ± 0.002 & 0.960 ± 0.000 \\
        \hline
        \multicolumn{5}{c}{Large Language Model} \\
        \hline
        Zero-shot & 0.826 ± 0.004 & 0.761 ± 0.009 & 0.950 ± 0.002 & 0.987 ± 0.001 \\
        Few-shot & 0.854 ± 0.002 & 0.792 ± 0.006 & 0.951 ± 0.001 & 0.981 ± 0.001 \\
        Chain-of-Thought & 0.827 ± 0.005 & 0.753 ± 0.003 & 0.940 ± 0.002 & 0.978 ± 0.001 \\
        Self-Consistency & 0.824 ± 0.002 & 0.760 ± 0.008 & 0.952 ± 0.001 & 0.987 ± 0.001 \\
        Least-to-Most & 0.832 ± 0.003 & 0.766 ± 0.007 & 0.944 ± 0.002 & 0.979 ± 0.001 \\
        Self-Reflection & 0.847 ± 0.001 & 0.796 ± 0.007 & 0.941 ± 0.001 & 0.976 ± 0.001 \\
        \hline
    \end{tabular}
\end{table}

For 14-region prediction, among neural models, XLM-RoBERTa achieved the highest performance with Accuracy of 0.690 and Macro-F1 of 0.631. CANINE and SVM also showed comparable performance (Accuracy of 0.683 and 0.682, respectively). It is noteworthy that while SVM achieved the highest performance in 99 nationality prediction, pre-trained models outperformed it in 14-region prediction. This suggests that pre-trained models possess extensive knowledge about languages and cultural spheres, and this knowledge functions effectively in region-level classification.

Among LLMs, Self-Reflection achieved the highest performance with Accuracy of 0.751 and Macro-F1 of 0.704. Zero-shot and Self-Consistency also showed high performance at 0.715 and 0.710, respectively. Similar to 99 nationality prediction, Least-to-Most remained at a relatively lower performance of 0.661. The difference between the best LLM and neural model performance was 0.061 (Self-Reflection 0.751 vs. XLM-RoBERTa 0.690), which is substantially reduced compared to the difference in 99 nationality prediction (0.295).

For 6-region prediction, overall performance improved. Among neural models, XLM-RoBERTa achieved Accuracy of 0.780 and Macro-F1 of 0.692, while CANINE showed comparable performance at 0.776. SVM achieved 0.767, with the gap from pre-trained models further widening. Among LLMs, Few-shot achieved the highest performance with Accuracy of 0.854 and Macro-F1 of 0.792. This contrasts with 14-region prediction where Few-shot performed below Zero-shot. At the coarser granularity of 6 regions, pattern presentation through examples appears to have functioned effectively. Self-Reflection also maintained high performance with Accuracy of 0.847.

Particularly noteworthy is the performance recovery of Least-to-Most. While it achieved Accuracy of 0.661 in 14-region prediction, substantially below other LLM methods, it reached 0.832 in 6-region prediction, narrowing the gap with other methods. Since Least-to-Most performs hierarchical inference (6 continents → 14 regions → 99 nationalities), when the final output is at the 6-region level, it is less susceptible to error propagation, which explains this result.

Analyzing performance changes across granularities reveals interesting trends. In the transition from 99 nationalities to 14 regions, the performance improvement of LLMs (Zero-shot: 0.759 → 0.715, a decrease of 0.044) was substantially smaller than that of neural models (SVM: 0.481 → 0.682, an increase of 0.201). While this may appear contradictory at first glance, considering that LLMs already achieved high performance in 99 nationality prediction, it can be interpreted that ceiling effects limited the room for improvement. In the transition from 14 regions to 6 regions, performance improved for all methods, with neural models showing particularly notable improvement.

Focusing on Precision@3, for 6-region prediction, LLMs achieved 0.976 or higher, and even neural models achieved 0.895 or higher, indicating an extremely high probability that the correct answer is included in the Top-3 predictions. This demonstrates that at the coarse granularity of 6 regions, models can capture the general geographical direction with high accuracy.

These results demonstrate a clear relationship between prediction granularity and performance differences between models. As granularity becomes coarser, the gap between LLMs and neural models narrows, reducing to 0.074 in Accuracy for 6-region prediction (Few-shot 0.854 vs. XLM-RoBERTa 0.780). This suggests that while character patterns alone provide sufficient information for coarse-grained prediction, the world knowledge possessed by LLMs provides a significant advantage for fine-grained prediction.

\subsection{Evaluation on Robustness to Label Frequency}

\begin{figure}[t]
    \centering
    \includegraphics[width=\linewidth]{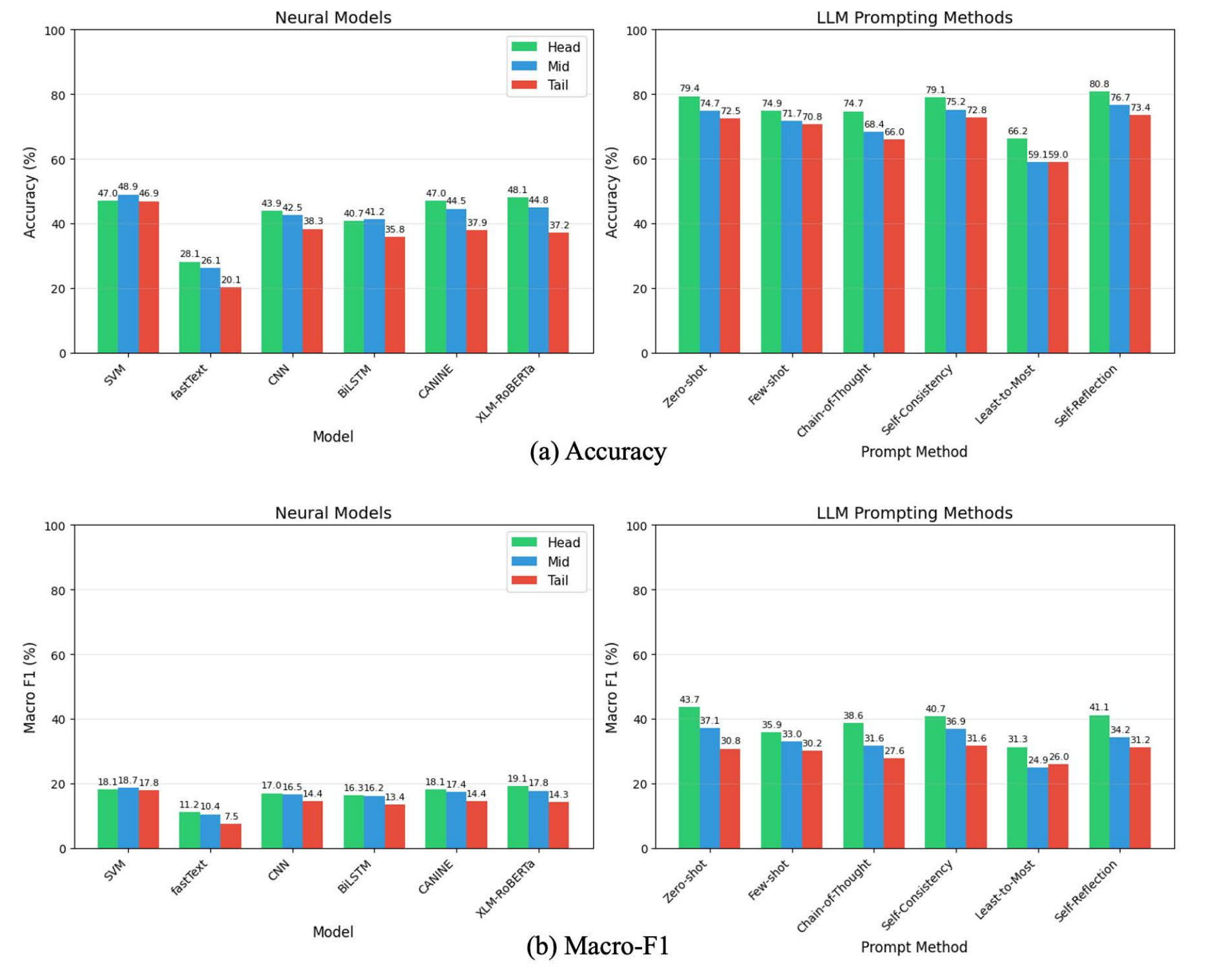}
    \caption{Accuracy and Macro-F1 by nationality frequency bin. Head, Mid, and Tail represent high-frequency, medium-frequency, and low-frequency nationality groups, respectively.}
    \label{fig:frequency_analysis}
\end{figure}

This section analyzes model robustness to nationality frequency. As described in Section 4.1.1, the 99 nationalities were equally divided into three groups based on their frequency in the training data: Head (high-frequency), Mid (medium-frequency), and Tail (low-frequency), and performance was evaluated for each group. Fig.~\ref{fig:frequency_analysis} shows the results for Accuracy and Macro-F1. Table~\ref{tab:head_tail_gap} shows the performance gap between Head and Tail groups ($\Delta$(H-T)) and the performance drop rate from Head (Drop\%) for each model. A larger $\Delta$(H-T) indicates higher frequency dependence, meaning that prediction of low-frequency nationalities is relatively more difficult.

Table~\ref{tab:head_tail_gap} shows the performance gap between Head and Tail groups ($\Delta$(H-T)) and the performance drop rate from Head (Drop\%) for each model. A larger $\Delta$(H-T) indicates higher frequency dependence, meaning that prediction of low-frequency nationalities is relatively more difficult.

\begin{table}[t]
\centering
\caption{Head vs. Tail performance gap (Accuracy).}
\label{tab:head_tail_gap}
\begin{tabular}{lccccc}
\hline
Model & Head & Mid & Tail & $\Delta$(H-T) & Drop\% \\
\hline
\multicolumn{6}{c}{Neural Model} \\
\hline
SVM & 0.470 & 0.489 & 0.469 & +0.001 & 0.3\% \\
fastText & 0.281 & 0.261 & 0.201 & +0.080 & 28.4\% \\
CNN & 0.439 & 0.425 & 0.383 & +0.057 & 12.9\% \\
BiLSTM & 0.407 & 0.412 & 0.359 & +0.049 & 12.0\% \\
CANINE & 0.471 & 0.446 & 0.379 & +0.092 & 19.5\% \\
XLM-RoBERTa & 0.481 & 0.449 & 0.372 & +0.109 & 22.6\% \\
\hline
\multicolumn{6}{c}{Large Language Model} \\
\hline
Zero-shot & 0.794 & 0.747 & 0.725 & +0.069 & 8.7\% \\
Few-shot & 0.749 & 0.717 & 0.708 & +0.042 & 5.5\% \\
Chain-of-Thought & 0.747 & 0.684 & 0.660 & +0.087 & 11.7\% \\
Self-Consistency & 0.791 & 0.752 & 0.728 & +0.063 & 8.0\% \\
Least-to-Most & 0.663 & 0.591 & 0.590 & +0.073 & 11.0\% \\
Self-Reflection & 0.808 & 0.767 & 0.734 & +0.074 & 9.2\% \\
\hline
\end{tabular}
\end{table}

Among neural models, pre-trained models exhibited the most pronounced frequency dependence. XLM-RoBERTa showed $\Delta$(H-T) of 0.109 and CANINE showed 0.092, with performance in the Tail group decreasing by approximately 20\% compared to the Head group. This suggests that pre-trained models are influenced by the frequency of languages and nationalities in the pre-training corpus. Since pre-training corpora contain more text in English and European languages, name patterns from these regions may be better learned.

In contrast, SVM showed $\Delta$(H-T) of only 0.001, demonstrating extremely high robustness. Since SVM is based on character n-gram pattern matching, it directly reflects the frequency distribution of training data, making it less prone to bias toward specific nationalities. fastText showed $\Delta$(H-T) of 0.080, but its Drop\% was the highest at 28.4\%, indicating that its already low performance deteriorated further in the Tail group.

For LLMs, $\Delta$(H-T) ranged from 0.042 to 0.087 across all methods. Few-shot was the most robust, with $\Delta$(H-T) of 0.042 and Drop\% of only 5.5\%. Since Few-shot selects examples from each region, examples from regions containing low-frequency nationalities may have had an effect of mitigating bias. On the other hand, Chain-of-Thought showed a relatively high $\Delta$(H-T) of 0.087, suggesting that biased judgments toward high-frequency nationalities are more likely to occur during the reasoning process.

Focusing on Macro-F1, the tendency of frequency dependence became more pronounced. LLM Macro-F1 ranged from approximately 0.350 to 0.440 in the Head group but decreased to 0.260 to 0.310 in the Tail group. For neural models, it ranged from approximately 0.130 to 0.190 in the Head group and 0.070 to 0.180 in the Tail group. Since Macro-F1 weights each class equally, performance degradation for individual nationalities in the Tail group is directly reflected. While LLMs show high robustness in Accuracy, performance degradation for low-frequency nationalities was clearly observed in Macro-F1.

From these results, the following insights were obtained regarding model frequency robustness. First, pre-trained models (CANINE, XLM-RoBERTa) inherit biases from pre-training corpora, making prediction of low-frequency nationalities relatively difficult. Second, simple methods such as SVM are the most robust to frequency. Third, while LLMs are superior in absolute performance, performance gaps still exist between high-frequency and low-frequency nationalities. Fourth, evaluation using Macro-F1 is important for revealing frequency biases that are difficult to observe with Accuracy alone.

\subsection{Error Analysis}
This section analyzes representative error patterns of LLM (Zero-shot) and neural model (XLM-RoBERTa), revealing qualitative differences in how the two approaches make errors.

First, we analyzed whether errors in nationality prediction are also errors at the region level, or become correct at the region level. Table~\ref{tab:region_accuracy} shows the results of region-level accuracy analysis. Even when LLM made incorrect nationality predictions, it predicted the correct region in 0.862 of cases. In contrast, XLM-RoBERTa's region accuracy remained at 0.668. This difference indicates that LLM possesses extensive knowledge about the relationships between names and regions, capturing the general direction even when making fine-grained errors.

\begin{table}[t]
\centering
\caption{Region-level accuracy analysis.}
\label{tab:region_accuracy}
\begin{tabular}{lcc}
\hline
Category & LLM (Zero-shot) & XLM-RoBERTa \\
\hline
Nationality Correct & 5,703 (0.757) & 3,292 (0.437) \\
Nationality Wrong, Region Correct & 788 (0.105) & 1,743 (0.231) \\
Nationality Wrong, Region Wrong & 1,043 (0.138) & 2,499 (0.332) \\
\hline
Region Accuracy (incl. Nationality Correct) & 0.862 & 0.668 \\
\hline
\end{tabular}
\end{table}

\begin{figure}[t]
    \centering
    \includegraphics[width=0.9\linewidth]{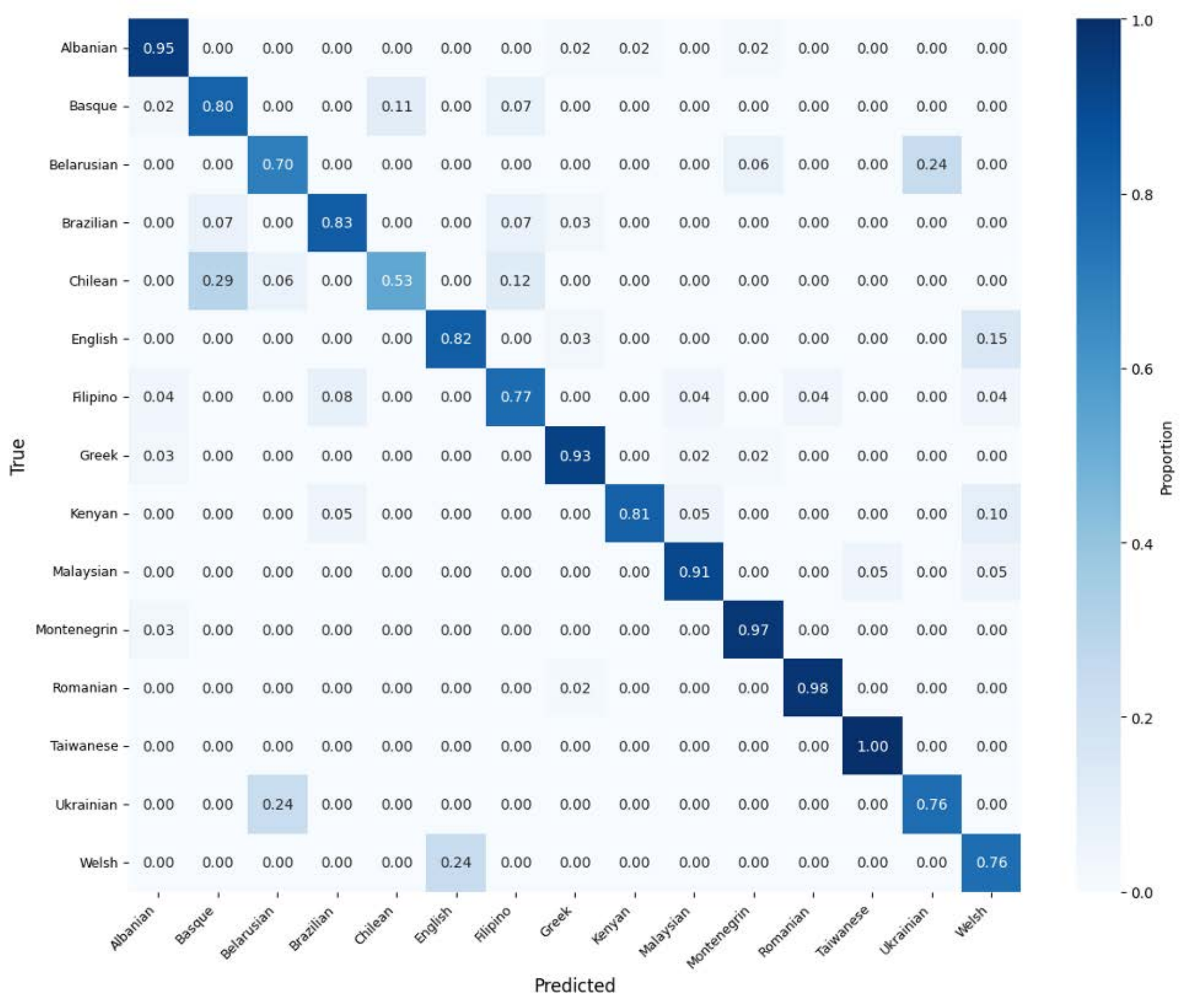}
    \caption{Confusion matrix for top 15 nationalities (XLM-RoBERTa).}
    \label{fig:confusion_neural}
\end{figure}

\begin{figure}[t]
    \centering
    \includegraphics[width=0.9\linewidth]{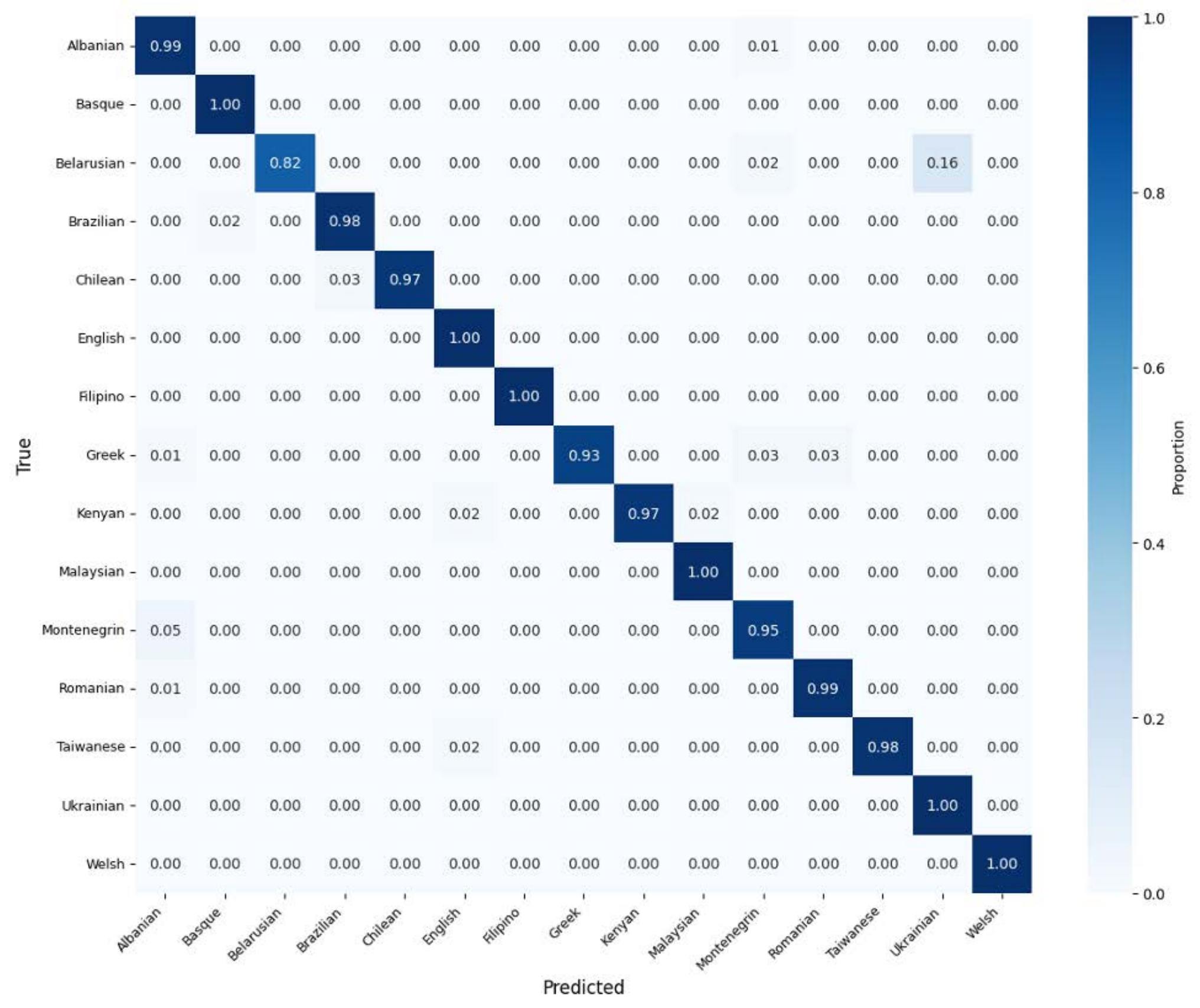}
    \caption{Confusion matrix for top 15 nationalities (GPT-4.1-mini Zero-shot).}
    \label{fig:confusion_llm}
\end{figure}

Fig.~\ref{fig:confusion_neural} and Fig.~\ref{fig:confusion_llm} show the confusion matrices for the top 15 nationalities for LLM (Zero-shot) and XLM-RoBERTa, respectively. The confusion matrix for LLM shows extremely high diagonal components ranging from 0.82 to 1.00, indicating few prediction errors. A minor error observed is the misclassification of Belarusian as Ukrainian (0.16), but this is an error within the same Eastern Europe region. On the other hand, the confusion matrix for XLM-RoBERTa shows high variance in diagonal components ranging from 0.53 to 1.0. Particularly low accuracy rates are observed for Chilean (0.53), Welsh (0.76), and Filipino (0.77), with off-diagonal values widely distributed. 29\% of Chilean samples were misclassified as Basque and 12\% as Filipino, indicating frequent cross-regional errors.

Table~\ref{tab:confusion_pairs} shows the top-10 confusion pairs and region agreement rates for both models. 0.667 of LLM's confusion pairs are errors within the same region. The most frequent confusion is English → British (67 cases), which is an error between extremely close nationalities within the same Western Europe region. Similarly, confusions between linguistically and culturally proximate nationalities such as Tamil → Indian, Welsh → British, and Uruguayan → Argentine are common. These errors are considered cases where distinction is inherently difficult due to extremely similar name patterns.

\begin{table}[t]
    \centering
    \caption{Top-10 confusion pairs.}
    \label{tab:confusion_pairs}
    \begin{tabular}{lcclcc}
        \hline
        \multicolumn{3}{c}{LLM (Zero-shot)} & \multicolumn{3}{c}{XLM-RoBERTa} \\
        True → Pred & Count & Same & True → Pred & Count & Same \\
        \hline
        English → British & 67 & \checkmark & Belarusian → Russian & 25 & \checkmark \\
        Tamil → Indian & 60 & \checkmark & Moldovan → Romanian & 22 & \checkmark \\
        Welsh → British & 46 & \checkmark & Brazilian → Portuguese & 21 & $\times$ \\
        Uruguayan → Argentine & 29 & \checkmark & Czech → Slovak & 21 & \checkmark \\
        Belarusian → Russian & 25 & \checkmark & Peruvian → Mexican & 20 & $\times$ \\
        Cuban → Mexican & 24 & $\times$ & Austrian → German & 20 & \checkmark \\
        Jamaican → American & 24 & $\times$ & Tunisian → Moroccan & 20 & \checkmark \\
        Taiwanese → Chinese & 22 & \checkmark & German → Austrian & 20 & \checkmark \\
        Australian → American & 22 & $\times$ & Kenyan → Ugandan & 19 & \checkmark \\
        Paraguayan → Argentine & 22 & \checkmark & Ecuadorian → Mexican & 19 & $\times$ \\
        \hline
        Region Agreement & \multicolumn{2}{c}{0.667} & Region Agreement & \multicolumn{2}{c}{0.600} \\
        \hline
    \end{tabular}
\end{table}

For XLM-RoBERTa's confusion pairs, the region agreement rate is 0.600, lower than LLM. A distinctive characteristic is the tendency for South American nationalities (Peruvian, Ecuadorian, Colombian, Guatemalan, Cuban) to be misclassified as Mexican. This is likely due to the high frequency of Mexican in the training data, resulting in prediction bias toward Mexican for Spanish-speaking names. Cross-regional errors such as Brazilian → Portuguese are also observed, indicating misclassification based on linguistic similarity (Portuguese language).

Table~\ref{tab:error_examples} shows representative examples of error patterns for both models. An interesting case is where LLM makes errors while the neural model is correct. For ``Jordan Williams'' (Welsh), LLM predicted American while XLM-RoBERTa correctly predicted Welsh. ``Jordan'' and ``Williams'' are names widely used in English-speaking countries, and LLM likely predicted American as the most common nationality. On the other hand, XLM-RoBERTa captured the characteristics of Welsh from training data patterns. Similarly, for ``Francis Zé'' (Cameroonian), LLM predicted French, suggesting that while it recognized French-language name patterns, it did not consider French-speaking African countries.

\begin{table}[t]
    \centering
    \caption{Representative error cases.}
    \label{tab:error_examples}
    \begin{tabular}{llccc}
        \hline
        Case Type & Name & True & LLM & Neural \\
        \hline
        Both Wrong, LLM Region \checkmark & José Manuel Hernández & Guatemalan & Mexican & Dominican \\
        Both Wrong, LLM Region \checkmark & Santokh Singh Matharu & Kenyan & Indian & Nepalese \\
        LLM Wrong, Neural Correct & Jordan Williams & Welsh & American & Welsh \\
        LLM Wrong, Neural Correct & Francis Zé & Cameroonian & French & Cameroonian \\
        LLM Correct, Neural Wrong & Dumitru Șchiopu & Romanian & Romanian & Moldovan \\
        LLM Correct, Neural Wrong & Everton Santos Bezerra & Brazilian & Brazilian & Mexican \\
        \hline
    \end{tabular}
\end{table}

Conversely, there are many examples where LLM is correct while the neural model makes errors. For ``Dumitru Șchiopu'' (Romanian), XLM-RoBERTa predicted Moldovan while LLM correctly predicted Romanian. For ``Everton Santos Bezerra'' (Brazilian), XLM-RoBERTa predicted Mexican, confirming the aforementioned bias toward Mexican.

From these analyses, it became clear that there are qualitative differences in the errors made by LLM and neural models. Since LLM makes predictions based on extensive world knowledge, it tends to misclassify to geographically and culturally close nationalities even when making errors. On the other hand, since neural models depend on training data patterns, they are prone to bias toward high-frequency classes and errors based on superficial similarity of character patterns. LLM errors tend to be ``near-miss'' errors, while neural model errors tend to be relatively more ``off-target'' errors that cross regional boundaries.

\section{Discussion}
\subsection{Key Findings}
In this study, we conducted a multifaceted comparison of neural models and LLMs on the nationality prediction task. The main results are as follows: i) LLMs significantly outperformed neural models in 99 nationality prediction (Self-Reflection: 0.776 vs. SVM: 0.481), ii) the performance gap between the two narrowed as prediction granularity became coarser (0.074 for 6-region prediction), iii) SVM showed the highest frequency robustness while pre-trained models exhibited notable performance degradation for low-frequency nationalities, iv) LLMs tended to make ``near-miss'' errors, predicting the correct region in 86.2\% of cases even when the nationality was incorrect, and v) advanced prompting methods except Self-Reflection and Self-Consistency performed below Zero-shot. Below, we discuss new insights derived from these results and implications for future research.

The results of this study suggest that LLMs' nationality prediction capability is based on world knowledge acquired during pre-training rather than fitting to training data. While neural models learn statistical correspondences between character patterns and nationality labels, LLMs are considered to leverage structured knowledge about the relationships between personal names and nationality, culture, language, and geography. This interpretation is consistent with the fact that LLMs are particularly advantageous in fine-grained prediction and tend to predict the correct region even when making errors. LLMs appear to make predictions by understanding higher-level concepts such as ``which cultural sphere does this name belong to,'' performing inference beyond mere pattern matching.

The relationship between granularity and performance gap reveals differences in the inherent difficulty of tasks and the capabilities required from models. At coarse granularity such as 6-region prediction, information obtained from character patterns (e.g., ``-ovic'' indicates Slavic, ``-ez'' indicates Spanish-speaking regions) is sufficient to achieve high performance. On the other hand, for 99 nationality prediction, more detailed cultural and linguistic knowledge is required to distinguish nationalities within the same region. For example, distinguishing Belarusian from Ukrainian requires understanding subtle differences in naming conventions between the two countries. This insight provides guidance for model selection according to the required granularity in practical applications. When coarse granularity is sufficient, computationally inexpensive neural models are effective, while LLMs are recommended when fine granularity is required.

The analysis of frequency robustness provides important insights regarding biases in pre-trained models. The fact that XLM-RoBERTa and CANINE showed performance degradation for low-frequency nationalities while SVM showed extremely high robustness suggests that the composition of pre-training corpora directly affects model prediction biases. Since pre-training corpora contain more text in English and European languages, name patterns from these regions may be better learned. This issue is shared by LLMs, with notable performance degradation for low-frequency nationalities observed in Macro-F1 evaluation. From a fairness perspective, ensuring diversity in pre-training data and developing bias correction methods during inference are important research challenges for the future.

The qualitative differences in errors between LLMs and neural models suggest the possibility of complementary utilization of both approaches. LLMs tend to make ``near-miss'' errors based on world knowledge, but conversely may overlook local patterns in training data. As shown in the error analysis, while LLM predicted American for ``Jordan Williams'' (Welsh), XLM-RoBERTa correctly predicted Welsh. This indicates that neural models capture specific patterns in training data. Ensemble methods that combine predictions from both models, such as a hybrid approach that references neural model predictions when LLM confidence is low, represent a promising direction that leverages the strengths of both approaches.

The results for prompting methods provide important insights regarding the relationship between task characteristics and prompt design. The fact that Few-shot and Chain-of-Thought performed below Zero-shot is attributed to the characteristics of the nationality prediction task. In Few-shot, limited examples may have caused excessive dependence on specific name patterns, hindering generalization to diverse names. In Chain-of-Thought, making the reasoning process explicit may have increased the risk of fixating on incorrect hypotheses (e.g., ``this suffix is characteristic of ~ language''). On the other hand, the effectiveness of Self-Reflection and Self-Consistency indicates that self-correction and integration of multiple inferences enhance prediction reliability. This insight suggests that task characteristics need to be considered when selecting prompting methods, providing important guidance for future prompt engineering research.

\subsection{Limitations}
This study has several limitations. First, LLM evaluation is limited to a single model, GPT-4.1-mini. Other LLMs such as Claude, Gemini, and Llama may exhibit different performance characteristics and bias patterns. In particular, how differences in pre-training data composition and training methods affect nationality prediction performance is an important research question that can be clarified by comparing multiple LLMs. Whether the findings of this study are applicable to LLMs in general requires future verification.

Second, the evaluation targets are limited to English (romanized) names. The name2nat dataset targets romanized names, and evaluation using original scripts such as Chinese characters, Arabic script, and Cyrillic script has not been conducted. In original scripts, the characters themselves provide strong clues about nationality and linguistic sphere, so performance differences and error patterns between models may differ. Additionally, the same name may have different representations depending on romanization methods, and the effect of notation normalization on prediction performance also warrants investigation.

Third, this study treats the correspondence between names and nationalities as static. In the real world, the relationship between names and nationalities changes dynamically due to immigration, naturalization, and international marriages. Furthermore, in cases where the same name is used across multiple nationalities, this dataset assigns only a single correct label. Evaluation frameworks that consider such inherent ambiguity, such as evaluation allowing multiple correct answers or prediction evaluation as probability distributions, are challenges for future research.

Despite these limitations, this study represents the first attempt to systematically compare the nationality prediction capabilities of LLMs and neural models, and the findings obtained provide a foundation for future research.

\section{Conclusion}
Personal names serve as an important source of information reflecting cultural and geographical background, and the task of predicting nationality and region from names holds practical value across various application domains. However, conventional approaches based on neural models learn correspondences between names and nationalities only from task-specific training data, posing challenges in generalizing to low-frequency nationalities and distinguishing similar nationalities within the same region. In this study, we conducted a multifaceted comparison of neural models and LLMs on the nationality and region prediction task. For neural models, we evaluated six types ranging from traditional machine learning methods to pre-trained language models, and for LLMs, we evaluated six prompting strategies ranging from simple zero-shot to self-correction methods. In addition to evaluation at three levels of granularity (nationality, region, and continent), we conducted frequency-based stratified analysis and error analysis.

Experimental results demonstrated that LLMs outperformed neural models across all prediction granularities, with the gap being particularly pronounced in fine-grained nationality prediction, while the performance gap narrowed as granularity became coarser. Frequency robustness analysis revealed that simple machine learning methods exhibited the highest robustness, while pre-trained models and LLMs showed performance degradation for low-frequency nationalities. Error analysis revealed that LLMs tend to make ``near-miss'' errors, predicting the correct region even when the nationality is incorrect, whereas neural models exhibited more pronounced cross-regional errors and bias toward high-frequency classes. These results indicate that the superiority of LLMs stems from world knowledge acquired during pre-training, that model selection based on required task granularity is important, and that evaluation should consider not only accuracy but also the quality of errors.

Future research directions include generalizing findings through comparison using multiple LLMs, evaluation across diverse writing systems including original scripts, development of hybrid methods that leverage the complementary characteristics of LLMs and neural models, and investigation of prompt design methods to mitigate frequency bias in LLMs.

\backmatter

\bmhead{Supplementary information}
Not applicable.

\bmhead{Acknowledgements}
The authors would like to thank The Nippon Foundation HUMAI Program for supporting this study and for providing a research environment that enabled the completion of this work. Regional map images used in figures are sourced from Wikimedia Commons and are licensed under Creative Commons (CC BY-SA 3.0/4.0).

\section*{Declarations}

\begin{itemize}
\item \textbf{Funding:} This work was supported by The Nippon Foundation HUMAI Program.
\item \textbf{Conflict of interest:} The author has no conflicts of interest to declare.
\item \textbf{Ethics approval:} Not applicable.
\item \textbf{Consent to participate:} Not applicable.
\item \textbf{Consent for publication:} Not applicable.
\item \textbf{Data availability:} The name2nat dataset used in this study is publicly available at \url{https://github.com/Kyubyong/name2nat}.
\item \textbf{Code availability:} Code will be made available upon reasonable request.
\item \textbf{Author contribution:} The author conducted all experiments, performed the analysis, and wrote the manuscript.
\end{itemize}









\bibliography{sn-bibliography}

\end{document}